\pdfoutput=1

\documentclass[11pt]{article}

\usepackage{acl}

\usepackage{amsmath}
\usepackage{amsfonts}
\usepackage{amssymb}
\usepackage{enumitem}
\usepackage{algorithm}
\usepackage{algpseudocode}
\usepackage{times}
\usepackage{latexsym}
\usepackage{soul}
\usepackage[T1]{fontenc}
\usepackage{verbatim}
\usepackage{booktabs}
\usepackage{hyperref}
\usepackage{multirow}
\usepackage[table]{colortbl}
\usepackage{graphicx}
\usepackage{amsmath}
\usepackage{caption}
\usepackage{subcaption}
\usepackage{tabularx}
\usepackage{url}
\usepackage{xurl}
\usepackage[most]{tcolorbox}
\tcbuselibrary{theorems}

\definecolor{LavenderBlush}{rgb}{1.0, 0.94, 0.96}
\definecolor{Lavender}{rgb}{0.9, 0.9, 0.98}
\definecolor{MistyRose}{rgb}{1.0, 0.89, 0.88}
\definecolor{MintCream}{rgb}{0.96, 1.0, 0.98}
\definecolor{AliceBlue}{rgb}{0.94, 0.97, 1.0}
\definecolor{Seashell}{rgb}{1.0, 0.96, 0.93}
\definecolor{LightYellow}{rgb}{1.0, 1.0, 0.88}
\definecolor{Peach}{rgb}{1.0, 0.9, 0.71}
\definecolor{Apricot}{rgb}{0.98, 0.81, 0.69}
\definecolor{LightCyan}{rgb}{0.88, 1.0, 1.0}

\definecolor{PeachPuff}{rgb}{1.0, 0.85, 0.73}
\definecolor{Beige}{rgb}{0.96, 0.96, 0.86}
\definecolor{LightSalmon}{rgb}{1.0, 0.63, 0.48}
\definecolor{Ivory}{rgb}{1.0, 1.0, 0.94}
\definecolor{MintCream}{rgb}{0.96, 1.0, 0.98}

\definecolor{Red}{rgb}{1.0, 0.0, 0.0}
\definecolor{Green}{rgb}{0.0, 0.5, 0.0}
\definecolor{Blue}{rgb}{0.0, 0.0, 1.0}
\definecolor{Orange}{rgb}{1.0, 0.55, 0.0}
\definecolor{Purple}{rgb}{0.5, 0.0, 0.5}
\definecolor{Goldenrod}{rgb}{0.85, 0.65, 0.13}
\definecolor{BurntOrange}{HTML}{CC5500}
\definecolor{Cyan}{rgb}{0.0, 1.0, 1.0}
\definecolor{Maroon}{rgb}{0.5, 0.0, 0.0}  

\definecolor{textblue}{RGB}{25,25,112}      
\definecolor{textred}{RGB}{139,0,0}         

\definecolor{softgray}{RGB}{220,220,220}
\definecolor{softblue}{RGB}{230,245,255}
\definecolor{softred}{RGB}{255,235,238}
\definecolor{framegray}{RGB}{100,100,100}
\definecolor{frameblue}{RGB}{100,149,237}   
\definecolor{framered}{RGB}{220,20,60}      

\definecolor{LightGreen}{rgb}{0.88, 1.0, 0.88}
\definecolor{LightPink}{rgb}{1.0, 0.9, 0.9}
\definecolor{LemonChiffon}{rgb}{1.0, 0.98, 0.8}

\definecolor{Brown}{rgb}{0.65, 0.16, 0.16}
\definecolor{LightSalmon}{rgb}{1.0, 0.63, 0.48}  
\definecolor{Gray}{rgb}{0.5, 0.5, 0.5}           
\definecolor{DarkGreen}{rgb}{0.0, 0.39, 0.0}

\usepackage[utf8]{inputenc}

\usepackage{microtype}

\usepackage{inconsolata}
\usepackage{graphicx}
\usepackage{subcaption}
\usepackage{tikz}
\usetikzlibrary{shapes.geometric, arrows.meta}

\title{Revealing the Truth with ConLLM for Detecting Multi-Modal Deepfakes}

\author{
\textbf{Gautam Siddharth Kashyap}\textsuperscript{1},
\textbf{Harsh Joshi}\textsuperscript{2},
\textbf{Niharika Jain}\textsuperscript{3},
\textbf{Ebad Shabbir}\textsuperscript{4}\\[-0.4ex]
\textbf{Jiechao Gao}\textsuperscript{5}\thanks{Corresponding Author: jiechao@stanford.edu, usman.naseem@mq.edu.au},
\textbf{Nipun Joshi}\textsuperscript{6},
\textbf{Usman Naseem}\textsuperscript{7}\footnotemark[1] \\
\textsuperscript{1, 7}Macquarie University, Sydney, Australia \\[-0.4ex]
\textsuperscript{2}Bharati Vidyapeeth's College Of Engineering, New Delhi, India \\[-0.4ex]
\textsuperscript{3}Vivekananda Institute of Professional Studies (VIPS), New Delhi, India \\[-0.4ex]
\textsuperscript{4}DSEU-Okhla, New Delhi, India \\[-0.4ex]
\textsuperscript{5}Center for SDGC, Stanford University, California, USA \\[-0.4ex]
\textsuperscript{6}Cornell University, New York, USA \\[-0.4ex]
}

\begin{document}
\maketitle
\begin{abstract}

The rapid rise of deepfake technology poses a severe threat to social and political stability by enabling hyper-realistic synthetic media capable of manipulating public perception. However, existing detection methods struggle with two core limitations: (1) \textit{modality fragmentation}, which leads to poor generalization across diverse and adversarial deepfake modalities; and (2) \textit{shallow inter-modal reasoning}, resulting in limited detection of fine-grained semantic inconsistencies. To address these, we propose \textbf{ConLLM} (Contrastive Learning with Large Language Models), a hybrid framework for robust multimodal deepfake detection. \textbf{ConLLM} employs a two-stage architecture: stage 1 uses Pre-Trained Models (PTMs) to extract modality-specific embeddings; stage 2 aligns these embeddings via contrastive learning to mitigate \textit{modality fragmentation}, and refines them using LLM-based reasoning to address \textit{shallow inter-modal reasoning} by capturing semantic inconsistencies. \textbf{ConLLM} demonstrates strong performance across audio, video, and audio-visual modalities. It reduces audio deepfake EER by up to 50\%, improves video accuracy by up to 8\%, and achieves approximately 9\% accuracy gains in audio-visual tasks. Ablation studies confirm that PTM-based embeddings contribute 9\%–10\% consistent improvements across modalities. Our code and data is available at: \url{https://github.com/gskgautam/ConLLM/tree/main}

\end{abstract}

\tcbset{
  mybox/.style={
    colback=Apricot!20,
    colframe=Apricot!80,
    coltitle=BurntOrange!80,
    fonttitle=\bfseries,
    sharp corners,
    boxrule=0.7pt,
    enhanced,
    attach boxed title to top left={xshift=1mm,yshift=-1mm},
    boxed title style={size=small,colback=Peach!30, colframe=BurntOrange!80, sharp corners}
  },
  labelstyle/.style={
    colback=LightCyan!30,
    colframe=Cyan!60,
    sharp corners,
    boxrule=0.5pt,
    fontupper=\scriptsize\sffamily\bfseries,
    left=2pt,
    right=2pt,
    top=1.5pt,
    bottom=1.5pt,
    enhanced
  }
}

\begin{figure}[t!]
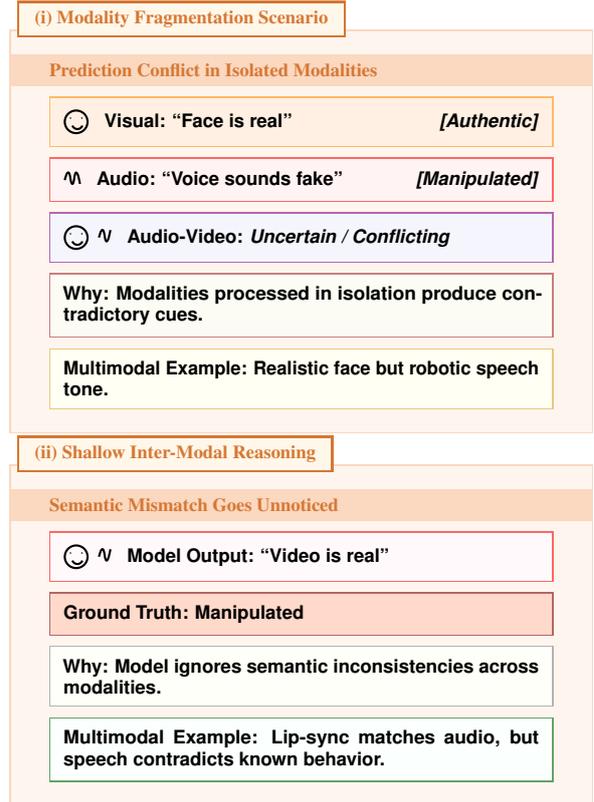

\centering
\scriptsize
\begin{minipage}{0.48\textwidth}
\begin{tcolorbox}[mybox,title={(i) Modality Fragmentation Scenario}]
\tcbsubtitle{Prediction Conflict in Isolated Modalities}

\begin{tcolorbox}[labelstyle, colback=PeachPuff!40, colframe=Orange!60]
\tikz[baseline=-0.5ex]{
  \draw[thick] (0,0) circle (0.15); 
  \fill (0.075,0.05) circle (0.01); 
  \fill (-0.075,0.05) circle (0.01); 
  \draw[thick] (-0.05,-0.05) arc[start angle=200,end angle=340,radius=0.07]; 
}
\hspace{2pt}
\textbf{Visual:} “Face is real” \hfill \textit{[Authentic]}
\end{tcolorbox}

\begin{tcolorbox}[labelstyle, colback=MistyRose!40, colframe=Red!60]
\tikz[baseline=-0.5ex]{
  \draw[thick] plot[smooth, tension=1] coordinates {(0,0) (0.05,0.1) (0.1,-0.05) (0.15,0.1) (0.2,-0.05)};
}
\hspace{2pt}
\textbf{Audio:} “Voice sounds fake” \hfill \textit{[Manipulated]}
\end{tcolorbox}

\begin{tcolorbox}[labelstyle, colback=Lavender!40, colframe=Purple!60]
\tikz[baseline=-0.5ex]{
  \draw[thick] (0,0) circle (0.15); \fill (0.075,0.05) circle (0.01); \fill (-0.075,0.05) circle (0.01);
  \draw[thick] (-0.05,-0.05) arc[start angle=200,end angle=340,radius=0.07];
  \draw[thick] plot[smooth, tension=1] coordinates {(0.3,0) (0.35,0.1) (0.4,-0.05) (0.45,0.1)};
}
\hspace{2pt}
\textbf{Audio-Video:} \textit{Uncertain / Conflicting}
\end{tcolorbox}

\begin{tcolorbox}[labelstyle, colback=Beige!30, colframe=Brown!60]
\textbf{Why:} Modalities processed in isolation produce contradictory cues.
\end{tcolorbox}

\begin{tcolorbox}[labelstyle, colback=LightYellow!40, colframe=Goldenrod!60]
\textbf{Multimodal Example:} Realistic face but robotic speech tone.
\end{tcolorbox}
\end{tcolorbox}
\end{minipage}
\hfill
\begin{minipage}{0.48\textwidth}
\begin{tcolorbox}[mybox,title={(ii) Shallow Inter-Modal Reasoning}]
\tcbsubtitle{Semantic Mismatch Goes Unnoticed}

\begin{tcolorbox}[labelstyle, colback=LavenderBlush!40, colframe=Red!60]
\tikz[baseline=-0.5ex]{
  \draw[thick] (0,0) circle (0.15); \fill (0.075,0.05) circle (0.01); \fill (-0.075,0.05) circle (0.01);
  \draw[thick] (-0.05,-0.05) arc[start angle=200,end angle=340,radius=0.07];
  \draw[thick] plot[smooth, tension=1] coordinates {(0.3,0) (0.35,0.1) (0.4,-0.05) (0.45,0.1)};
}
\hspace{2pt}
\textbf{Model Output:} “Video is real”
\end{tcolorbox}

\begin{tcolorbox}[labelstyle, colback=LightSalmon!40, colframe=Maroon!60]
\textbf{Ground Truth:} Manipulated
\end{tcolorbox}

\begin{tcolorbox}[labelstyle, colback=Ivory!40, colframe=Gray!60]
\textbf{Why:} Model ignores semantic inconsistencies across modalities.
\end{tcolorbox}

\begin{tcolorbox}[labelstyle, colback=MintCream!40, colframe=DarkGreen!60]
\textbf{Multimodal Example:} Lip-sync matches audio, but speech contradicts known behavior.
\end{tcolorbox}
\end{tcolorbox}
\end{minipage}

\caption{Illustration of common failure modes in multimodal deepfake detection. \textbf{Top:} Inconsistent unimodal predictions arise from poorly aligned audio-visual features (\textit{modality fragmentation}). \textbf{Bottom:} Surface-level fusion leads to incorrect predictions due to the model’s inability to reason over subtle cross-modal inconsistencies (\textit{shallow inter-modal reasoning}).}
\label{fig:limitations_final}
\end{figure}

\section{Introduction}
\label{sec:Introduction}

The rapid proliferation of deepfake technology, driven by advances in generative modeling and multimodal synthesis, poses growing societal and political risks~\cite{kashyap2025clarity, al2023impact}. High-profile incidents—such as synthetic videos portraying political figures in fabricated scenarios—have underscored the need for effective and generalizable detection systems~\cite{phukan2024paralinguistic, islam2024ai, agarwal2021detecting}. As deepfake techniques evolve, they increasingly blend audio, video, and textual modalities using sophisticated architectures, including GANs, VAEs, and transformer-based TTS systems~\cite{govindarajan2025magic, masood2023deepfakes, yang2023avoid, cao2022end}. These developments have made it significantly more difficult to distinguish between real and manipulated content.

Despite ongoing progress, existing detection methods predominantly focus on unimodal features, such as visual inconsistencies in facial movements~\cite{tan2019efficientnet, chollet2017xception} or artifacts in synthesized speech~\cite{phukan2024heterogeneity, ranjan2022statnet}. However, multimodal deepfakes exploit cross-modal coherence to create more convincing forgeries, rendering unimodal detectors insufficient~\cite{hao2022deepfake}. While some multimodal approaches exist~\cite{feng2023self, yu2023pvass}, they often suffer from two key limitations (see Figure \ref{fig:limitations_final}): (1) \textit{modality fragmentation}—where features extracted from different modalities remain isolated and poorly aligned, leading to poor generalization; and (2) \textit{shallow inter-modal reasoning}—the inability to effectively model complex semantic relationships and subtle inconsistencies across modalities, limiting detection accuracy. 

Therefore, to address the aforementioned challenges, we propose \textbf{ConLLM} (Contrastive Learning with Large Language Models), a hybrid framework that integrates contrastive learning and large-scale pretrained LLMs (i.e., GPT-style transformer architectures) for robust deepfake detection across audio, visual, and audio-visual modalities. The framework follows a two-stage modular architecture. In Stage 1, Pre-Trained Models (PTMs) independently extract embeddings from each modality, preserving modality-specific characteristics without premature fusion. These embeddings are then refined in Stage 2, where contrastive learning aligns semantically consistent representations and separates forgeries from authentic content, directly addressing \textit{modality fragmentation}. Simultaneously, LLM-based reasoning modules are applied to the aligned embeddings to capture subtle semantic inconsistencies and inter-modal dependencies, mitigating the limitations of \textit{shallow inter-modal reasoning}. \textbf{ConLLM} guarantees contextual robustness and discriminative power by separating feature extraction and semantic integration. In summary, our main contributions are as follows:
\begin{itemize}
    \item We propose \textbf{ConLLM}, a novel two-stage hybrid framework that first performs modality-specific embedding extraction via PTMs, followed by contrastive alignment and LLM-based reasoning for semantic refinement—addressing both \textit{modality fragmentation} and \textit{shallow inter-modal reasoning}.
    \vspace{-0.3cm}
    \item We comprehensively evaluate \textbf{ConLLM} across three modalities—audio, video, and audio-visual—to demonstrate its generalizability and robustness.
    \vspace{-0.3cm}
    \item \textbf{ConLLM} outperforms state-of-the-art baselines across all modalities, reducing audio deepfake EER by up to 50\%, improving video accuracy by up to 8\%, and achieving 9\% accuracy gains in audio-visual tasks. PTM-based embeddings further contribute 9\%–10\% consistent performance improvements in ablations.
\end{itemize}


\begin{figure*}[hbt!]
\centering
    \includegraphics[width=.95\linewidth]{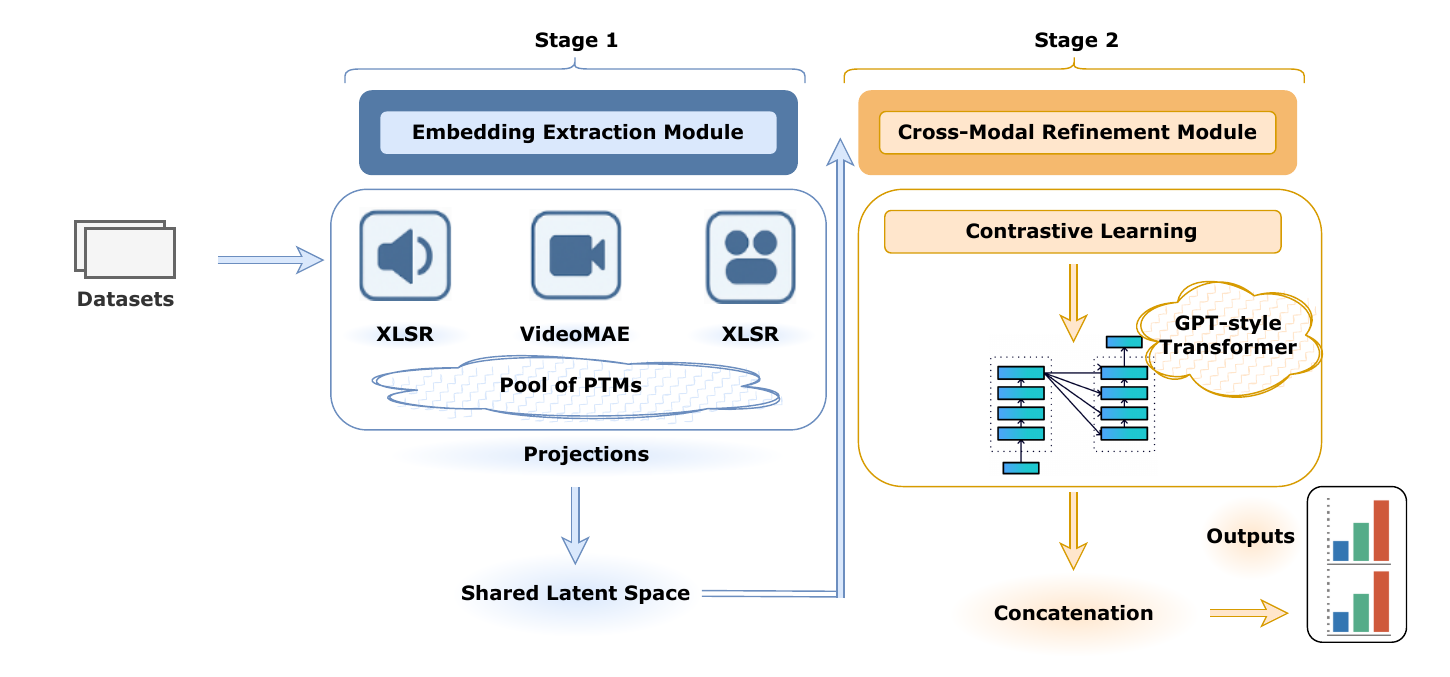}
    \caption{Overview of the \textbf{ConLLM} architecture. Modality-specific embeddings are extracted using PTMs (XLS-R, VideoMAE, VATLM) and projected into a shared latent space. A contrastive learning block aligns embeddings, followed by LLM-based refinement using a transformer. Refined embeddings are concatenated and passed through a classifier to detect deepfake content.}
    \label{ConLLM}
\end{figure*}

\section{Related Works}
\label{Related}

\paragraph{Contrastive Learning.}
Contrastive learning has been widely used to improve representation learning by encouraging semantically similar instances to be closer in the embedding space while pushing dissimilar ones apart \cite{zhai2022lit}. Recent works have explored contrastive objectives in unimodal settings such as voice spoofing \cite{kashyap2025fooling} and image forgery detection \cite{yu2024diffforensics}. However, these approaches often focus on single-modality embeddings and do not address cross-modal alignment, which is critical in multimodal deepfake detection.

\paragraph{Large Language Models.}
Large Language Models (LLMs), such as GPT-style transformers, have demonstrated impressive capabilities in semantic reasoning \cite{openai2023gpt4} and few-shot learning \cite{wei2022chain}. While LLMs have been used for vision-language tasks \cite{alayrac2022flamingo}, their role in deepfake detection remains underexplored. Most existing detection models \cite{qi2023deepfakebench} rely on shallow fusion or early fusion strategies without leveraging LLMs' capability for high-level reasoning. 

\paragraph{Multimodal Deepfake Detection.}
Multimodal approaches have emerged to address the limitations of unimodal detection by fusing features from audio and visual streams. Early works relied on handcrafted fusion strategies \cite{chugh2020not, mittal2020emotions}, while recent methods use neural architectures for cross-modal learning \cite{kashyap2025childguard, cheng2023voice, feng2023self, yu2023pvass, cai2022you}. Despite performance gains, these models often suffer from \textit{modality fragmentation} and \textit{shallow inter-modal reasoning}.

\section{Model Architecture} 
The proposed \textbf{ConLLM} framework utilizes a novel two-stage architecture, as illustrated in Figure~\ref{ConLLM}, to address the key challenges of \textit{modality fragmentation} and \textit{shallow inter-modal reasoning} in multimodal deepfake detection.

\subsection{Embedding Extraction Module}

This module extracts modality-specific embeddings from \textit{audio}, \textit{video}, and \textit{audio-visual} inputs using PTMs. For audio, XLS-R converts speech signals into robust representations; for video, VideoMAE captures spatiotemporal features; and for audio-visual data, VATLM jointly models both modalities to encode cross-modal interactions. Each PTM generates high-dimensional embeddings, which are then mapped to a shared latent space via modality-specific projection functions. These functions align heterogeneous embeddings into a unified space, enabling joint learning across modalities in later stages.

Formally, let $\mathcal{D} = \{\mathcal{D}_a, \mathcal{D}_v, \mathcal{D}_{av}\}$ denote the input dataset, where $\mathcal{D}_m$ corresponds to each modality $m \in \{a, v, av\}$. Each $\mathcal{D}_m$ is passed through a PTM $\phi_m$ to yield embeddings $\mathbf{z}_m = \phi_m(\mathcal{D}_m)$, where $\mathbf{z}_m \in \mathbb{R}^{d_m}$. These are then projected to a shared latent space $\mathbb{R}^{d_s}$ using $\psi_m$, yielding $\mathbf{h}_m = \psi_m(\mathbf{z}_m)$ with $\mathbf{h}_m \in \mathbb{R}^{d_s}$. This ensures embedding compatibility for subsequent alignment and reasoning.

\subsection{Cross-Modal Refinement Module}

The second module applies contrastive learning to align embeddings across modalities and utilizes a GPT-style transformer architecture to refine the learned multimodal representations. The objective is to minimize the distance between embeddings of authentic samples while maximizing the distance between manipulated ones.

Let $\mathcal{H} = \{\mathbf{h}_a, \mathbf{h}_v, \mathbf{h}_{av}\}$ represent the set of embeddings for a given data sample, where $\mathbf{h}_a$, $\mathbf{h}_v$, and $\mathbf{h}_{av}$ denote the audio, visual, and fused audio-visual embeddings, respectively. The contrastive loss $\mathcal{L}_{\text{contrast}}$ is computed as
$\mathcal{L}_{\text{contrast}} = -\sum_{(i, j) \in \mathcal{P}} \log \frac{\exp(\text{sim}(\mathbf{h}_i, \mathbf{h}_j) / \tau)}{\sum_{k \in \mathcal{N}} \exp(\text{sim}(\mathbf{h}_i, \mathbf{h}_k) / \tau)}$
where $\mathcal{P}$ denotes the set of positive pairs (e.g., authentic samples across modalities), $\mathcal{N}$ denotes negative pairs, $\text{sim}(\cdot,\cdot)$ computes cosine similarity, and $\tau$ is the temperature parameter.

After contrastive alignment, the embeddings $\mathbf{h}_m^{\text{input}} = \{\mathbf{h}_a, \mathbf{h}_v, \mathbf{h}_{av}\}$ are input to a GPT-style transformer module denoted as $\phi_{\text{Transformer}}$, which models inter-modal semantic dependencies and contextualizes the embeddings:
$\mathbf{h}_m^{\text{refined}} = \phi_{\text{Transformer}}(\mathbf{h}_m^{\text{input}}) = \mathcal{F}_{\text{Transformer}}(\mathbf{h}_a, \mathbf{h}_v, \mathbf{h}_{av})$.
The core of $\mathcal{F}_{\text{Transformer}}$ is the multi-head self-attention mechanism defined as
$\text{Attention}(\mathbf{Q}, \mathbf{K}, \mathbf{V}) = \text{softmax}\left(\frac{\mathbf{Q}\mathbf{K}^\top}{\sqrt{d_k}}\right) \mathbf{V}$
where the query $\mathbf{Q}$, key $\mathbf{K}$, and value $\mathbf{V}$ matrices are computed from the input embeddings using learned linear projections:
$\mathbf{Q} = \mathbf{W}_Q \mathbf{h}_m$, $\mathbf{K} = \mathbf{W}_K \mathbf{h}_m$, $\mathbf{V} = \mathbf{W}_V \mathbf{h}_m$
with $\mathbf{W}_Q$, $\mathbf{W}_K$, and $\mathbf{W}_V$ being trainable parameters, and $d_k$ the dimensionality of the keys.

The refined embeddings $\mathbf{h}_m^{\text{refined}}$ are concatenated into a single vector:
$\mathbf{h}_{\text{concat}} = [\mathbf{h}_a^{\text{refined}}, \mathbf{h}_v^{\text{refined}}, \mathbf{h}_{av}^{\text{refined}}]$
which is then passed through a classification head\footnote{The final classification is performed using a sigmoid activation function.} $f_{\text{cls}}$ to produce the final prediction:
$\hat{y} = f_{\text{cls}}(\mathbf{h}_{\text{concat}})$, where $\hat{y} \in \{0, 1\}$.

Our cross-modal refinement module is inspired by the transformer architecture popularized by GPT models, which use self-attention to capture complex dependencies. Unlike the official GPT models—which publicly support primarily text (and in some variants audio or video)—our architecture is customized to jointly process and reason over audio, visual, and audio-video embeddings.

\section{Experimental Setup} 

We select modality-specific PTMs based on the SUPERB benchmark~\cite{yang2021superb} to ensure high-quality feature extraction. For audio, we use \textbf{XLS-R}\footnote{\url{https://huggingface.co/facebook/wav2vec2-xls-r-1b}}~\cite{babu2021xls}, which achieves strong performance in speech tasks \cite{hsu2021hubert} and has shown promise in audio deepfake detection \cite{hsu2021hubert}. For video, \textbf{VideoMAE}\footnote{\url{https://huggingface.co/docs/transformers/en/model_doc/videomae}}~\cite{tong2022videomae} is used for its efficient spatiotemporal feature learning and success in video understanding \cite{arnab2023vivit}. For audio-visual inputs, we employ \textbf{VATLM}\footnote{\url{https://github.com/microsoft/SpeechT5/tree/main/VATLM}}~\cite{zhu2023vatlm}, a unified transformer effective in multimodal representation learning across speech and vision \cite{gong2022contrastive}.

\subsection{Datasets}
\label{Dataset}

We evaluate \textbf{ConLLM} on six widely-used benchmarks across three modalities. For audio-based detection, we use the \textbf{ASVSpoof 2019 (LA)}\footnote{\url{https://www.asvspoof.org/index2019.html}} dataset (ASV) \cite{wang2020asvspoof}, which provides both bona fide and spoofed speech samples. We follow the official LA protocol for training and evaluation. We also use \textbf{DECRO}\footnote{\url{https://zenodo.org/records/7603208}} \cite{ba2023transferring}, a multilingual dataset with English (\textbf{D-E}) and Chinese (\textbf{D-C}) audio samples to assess cross-lingual performance. 

For video-based deepfake detection, we use \textbf{Celeb-DF (CDF)}\footnote{\url{https://github.com/yuezunli/celeb-deepfakeforensics}} \cite{li2020celeb}, containing 590 real and 5,639 manipulated videos with enhanced synthesis quality, and \textbf{WildDeepfake (WD)}\footnote{\url{https://github.com/OpenTAI/wild-deepfake}} \cite{zi2020wilddeepfake}, which includes 707 deepfake videos from uncontrolled environments to evaluate real-world generalizability.

For audio-visual detection, we use \textbf{FakeAVCeleb (FAFC)}\footnote{\url{https://github.com/DASH-Lab/FakeAVCeleb}} \cite{khalid2021fakeavceleb}, which includes 500 real and 19,500 deepfake samples, and the \textbf{DeepFake Detection Challenge (DFDC)}\footnote{\url{https://ai.meta.com/datasets/dfdc/}} \cite{dolhansky2020deepfake}, the largest available benchmark with 128,154 manipulated videos across diverse subjects and environments.

\textbf{\textit{Note:}} We adhere to the official training and evaluation splits provided by each dataset’s authors.

\subsection{Evaluation Metrics}
\label{Evaluation}

We evaluate the performance of \textbf{ConLLM} using three standard metrics: Equal Error Rate (EER), Area Under the Curve (AUC), and Accuracy (ACC). These metrics are selected based on the modality and nature of the task.

For audio-based deepfake detection, we report \textbf{Equal Error Rate (EER)}, which is the point at which the False Positive Rate (FPR)\footnote{The proportion of negative samples incorrectly classified as positive.} equals the False Negative Rate (FNR)\footnote{The proportion of positive samples incorrectly classified as negative.}. EER provides a concise and meaningful indicator of verification accuracy, especially suited for binary spoof detection.

For video and audio-visual deepfake detection, we report both \textbf{Area Under the Curve (AUC)} and \textbf{Accuracy (ACC)}. AUC reflects the model’s ability to distinguish between authentic and manipulated content across different thresholds, while Accuracy measures the overall proportion of correctly classified samples.

\textbf{\textit{Note:}} In all result tables, \textuparrow~indicates that higher values are preferable, whereas \textdownarrow~indicates that lower values are preferable.


\subsection{Hyperparameters}

For training and evaluation, \textbf{ConLLM} was configured with the following hyperparameters: a learning rate of $10^{-3}$, batch size of 32, and a total of 50 epochs. Momentum was set to 0.9, and weight decay to $10^{-4}$. A dropout rate of 0.5 was used to mitigate overfitting, and ReLU was selected as the activation function. The model was optimized using the Adam optimizer. A step decay learning rate schedule was employed, with a decay factor of 0.1 applied every 10 epochs. \textit{\textbf{Note:}} All hyperparameters were held constant across datasets to ensure fair and consistent performance evaluation.

\section{Result Analysis}
\subsection{Comparison to State-of-the-Art}

Tables~\ref{tableEERComparison}, \ref{tableDeepfakeMethods}, and \ref{tableAudioVisualMethods} demonstrate the strong performance of \textbf{ConLLM} on deepfake detection across multiple modalities.

In audio deepfake detection (see Table~\ref{tableEERComparison}), \textbf{ConLLM} achieves exceptionally low Equal Error Rates (EER) of 0.21\% on the ASV dataset and 0.01\% on the D-E dataset. These results outperform prior works such as CL+GL \cite{kashyap2025fooling},  MiO~\cite{phukan2024heterogeneity} and Res-TSSDNet~\cite{ba2023transferring}. Note that since D-C was used primarily as a test set for cross-lingual evaluation in \cite{ba2023transferring}, we exclude models trained and tested on D-C from this comparison. This highlights \textbf{ConLLM}'s capability to detect subtle audio manipulations reliably.

For video deepfake detection (see Table~\ref{tableDeepfakeMethods}), \textbf{ConLLM} achieves accuracy rates of 98.75\% and 85.00\% on the CDF and WD datasets, respectively, with Area Under Curve (AUC) values of 99.98\% and 94.50\%. These results surpass existing approaches such as RECCE~\cite{cao2022end} and Multi-Attention~\cite{zhao2021multi}, demonstrating strong visual feature analysis.

In audio-visual tasks (see Table~\ref{tableAudioVisualMethods}), \textbf{ConLLM} achieves accuracies of 98.75\% on FAVC and 96.50\% on DFDC, along with AUC scores of 99.98\% and 98.71\%, respectively. This outperforms state-of-the-art models including PVASS~\cite{yu2023pvass} and AVAD~\cite{feng2023self}, showcasing effective multimodal fusion.

\begin{table}[t!]
\scriptsize
\renewcommand{\arraystretch}{1.1}
\centering
\setlength{\tabcolsep}{5pt}
\begin{tabular}{lcc}
\toprule
\textbf{Model} & \textbf{ASV~\textdownarrow} & \textbf{D-E~\textdownarrow} \\
\midrule
CQT-DCT-LCNN \cite{lavrentyeva2019stc} & 1.84 & -- \\
STATNet \cite{ranjan2022statnet}       & --   & 0.20 \\
Res-TSSDNet \cite{ba2023transferring}  & --   & 0.02 \\
MiO \cite{phukan2024heterogeneity}     & 0.41 & 0.04 \\
CL+GL \cite{kashyap2025fooling}        & 0.33 & 0.01 \\
\midrule
\textbf{ConLLM w/ GPT Weights}         & \textbf{0.21} & \textbf{0.01} \\
ConLLM w/o GPT Weights                 & 0.50 & 0.05 \\
\bottomrule
\end{tabular}
\caption{Performance comparison of EER (\%) for audio deepfake detection on ASVSpoof 2019 (LA) and DECRO datasets.}
\label{tableEERComparison}
\end{table}

\begin{table}[!htpb]
\scriptsize
\centering
\resizebox{\columnwidth}{!}{
\begin{tabularx}{\columnwidth}{l *{2}{>{\centering\arraybackslash}X} *{2}{>{\centering\arraybackslash}X}}
\toprule
\textbf{Method} & \textbf{CDF ACC (\%)~\textuparrow} & \textbf{CDF AUC (\%)~\textuparrow} & \textbf{WD ACC (\%)~\textuparrow} & \textbf{WD AUC (\%)~\textuparrow} \\
\midrule
Xception \cite{chollet2017xception} & 97.90 & 99.73 & 77.25 & 86.76 \\
EfficientNet-B4 \cite{tan2019efficientnet} & 97.63 & 99.20 & 81.63 & 90.36 \\
F3-Net \cite{qian2020thinking} & 95.95 & 98.93 & 80.66 & 87.53 \\
Mutil-Attention \cite{zhao2021multi} & 97.92 & 99.94 & 82.86 & 90.71 \\
RFM \cite{wang2021representative} & 97.96 & 99.94 & 77.38 & 83.92 \\
RECCE \cite{cao2022end} & 98.59 & 99.94 & 83.25 & 92.02 \\
\midrule
\textbf{ConLLM w/ GPT Weights} & \textbf{98.75} & \textbf{99.98} & \textbf{85.00} & \textbf{94.50} \\
ConLLM w/o GPT Weights & 96.50 & 99.20 & 81.00 & 90.00 \\
\bottomrule
\end{tabularx}
}
\caption{Performance Comparison of Video Deepfake Detection Models on Celeb-DF and WildDeepfake Datasets}
\label{tableDeepfakeMethods}
\end{table}

\begin{table}[t!]
\scriptsize
\centering
\resizebox{\columnwidth}{!}{
\begin{tabularx}{\columnwidth}{l *{4}{>{\centering\arraybackslash}X} *{4}{>{\centering\arraybackslash}X}}
\toprule
\textbf{Method} & \textbf{FAVC ACC (\%)~\textuparrow} & \textbf{FAVC AUC (\%)~\textuparrow} & \textbf{DFDC ACC (\%)~\textuparrow} & \textbf{DFDC AUC (\%)~\textuparrow} \\
\midrule
MDS \cite{chugh2020not} & 82.8 & 86.5 & 89.8 & 91.6 \\
EmoForen \cite{mittal2020emotions} & 78.1 & 79.8 & 80.6 & 84.4 \\
JointAV \cite{zhou2021joint} & 82.5 & 83.3 & 90.2 & 91.9 \\
BA-TFD \cite{cai2022you} & 80.8 & 84.9 & 79.1 & 84.6 \\
AVFakeNet \cite{ilyas2023avfakenet} & 78.4 & 83.4 & 82.8 & 86.2 \\
VFD \cite{cheng2023voice} & 81.5 & 86.1 & 80.9 & 85.1 \\
AVoiD-DF \cite{yang2023avoid} & 83.7 & 89.2 & 91.4 & 94.8 \\
AVAD \cite{feng2023self} & 94.2 & 94.5 & 93.2 & 96.7 \\
PVASS \cite{yu2023pvass} & 95.7 & 97.3 & 96.3 & 98.9 \\
\midrule
\textbf{ConLLM w/ GPT Weights} & \textbf{98.75} & \textbf{99.98} & \textbf{96.50} & \textbf{98.92} \\
ConLLM w/o GPT Weights & 95.00 & 98.00 & 90.00 & 94.00 \\
\bottomrule
\end{tabularx}
}
\caption{Performance Comparison of Audio-Visual Deepfake Detection on FakeAVCeleb and DeepFake Detection Challenge Datasets}
\label{tableAudioVisualMethods}
\end{table}

To evaluate the impact of large-scale pretrained language model architectures, we compare two versions of \textbf{ConLLM}--one initialized with pretrained transformer weights inspired by GPT-style architectures, and another trained from scratch without such initialization. The pretrained variant consistently outperforms the non-pretrained model, demonstrating the benefits of leveraging pretrained knowledge embedded in these transformer architectures, beyond merely increasing model capacity.

\subsection{Computational Experiments}
\label{Additional}

In this section, we evaluate the computational efficiency of \textbf{ConLLM} across multiple benchmark datasets using three key metrics: Memory Usage\footnote{The amount of RAM consumed by a model during inference.} (MU), Floating Point Operations\footnote{The total number of arithmetic computations performed by a model.} (FLOPs), and Inference Time\footnote{The time taken by a model to process an input and generate an output.} (IT). These metrics are essential for understanding the feasibility of deploying deepfake detection models in real-world scenarios.

\begin{table}[t!]
\centering
\scriptsize
\setlength{\tabcolsep}{4pt} 
\renewcommand{\arraystretch}{0.9} 
\begin{tabular}{@{}l|l|ccc@{}}
\toprule
\textbf{Dataset} & \textbf{Method} & \textbf{IT (ms) ↓} & \textbf{MU (GB) ↓} & \textbf{FLOPs ↓} \\
\midrule
\multicolumn{5}{c}{\textbf{Audio}} \\
\midrule
\multirow{4}{*}{ASV}  
  & Audio Flamingo    & 280 & 11.5 & 190 \\
  & Audio Flamingo 2  & 170 & 5.2  & 105 \\
  & Audio Flamingo 3  & 130 & 4.1  & 85  \\
  & \textbf{ConLLM}   & \textbf{55}  & \textbf{1.6}  & \textbf{45}  \\
\midrule
\multirow{4}{*}{D-E}  
  & Audio Flamingo    & 285 & 11.7 & 195 \\
  & Audio Flamingo 2  & 175 & 5.3  & 108 \\
  & Audio Flamingo 3  & 135 & 4.3  & 88  \\
  & \textbf{ConLLM}   & \textbf{60}  & \textbf{1.8}  & \textbf{48}  \\
\midrule
\multicolumn{5}{c}{\textbf{Video}} \\
\midrule
\multirow{3}{*}{CDF}  
  & Video-LLM         & 320 & 12.8 & 215 \\
  & MiniGPT4-Video    & 180 & 5.8  & 115 \\
  & \textbf{ConLLM}   & \textbf{65}  & \textbf{2.0}  & \textbf{55}  \\
\midrule
\multirow{3}{*}{WD}   
  & Video-LLM         & 330 & 13.1 & 220 \\
  & MiniGPT4-Video    & 185 & 6.0  & 118 \\
  & \textbf{ConLLM}   & \textbf{70}  & \textbf{2.2}  & \textbf{60}  \\
\midrule
\multicolumn{5}{c}{\textbf{Audio-Visual}} \\
\midrule
\multirow{3}{*}{FAFC} 
  & Video-LLaMA       & 360 & 13.8 & 235 \\
  & video-SALMONN-o1  & 190 & 6.3  & 125 \\
  & \textbf{ConLLM}   & \textbf{75}  & \textbf{2.5}  & \textbf{70}  \\
\midrule
\multirow{3}{*}{DFDC} 
  & Video-LLaMA       & 365 & 14.0 & 240 \\
  & video-SALMONN-o1  & 195 & 6.6  & 130 \\
  & \textbf{ConLLM}   & \textbf{80}  & \textbf{2.8}  & \textbf{75}  \\
\bottomrule
\end{tabular}
\caption{Computational efficiency of \textbf{ConLLM} compared to modality-specific large models. ConLLM achieves significantly reduced inference time (IT), memory usage (MU), and FLOPs across all modalities.}
\label{tab:performance_comparison}
\end{table}

Table~\ref{tab:performance_comparison} compares \textbf{ConLLM} with representative multimodal models, including AudioFlamingo \cite{kong2024audio}, AudioFlamingo 2 \cite{ghosh2025audio}, and AudioFlamingo 3 \cite{ghosh2025audio} for audio datasets, Video-LLM \cite{cheng2024videollama2} and MiniGPT4-Video \cite{ataallah2024minigpt4video} for video datasets, and Video-LLaMA \cite{zhang2023videollama} and video-SALMONN-o1 \cite{tang2025salmonn} for audio-visual datasets. The results demonstrate that \textbf{ConLLM} consistently achieves superior efficiency. For example, on the ASV dataset, \textbf{ConLLM} processes inputs in approximately 55 ms using 1.6 GB of memory and 45 GFLOPs—significantly outperforming AudioFlamingo 3 \cite{ghosh2025audio}, which requires around 130 ms, 4.1 GB, and 85 GFLOPs, respectively.

This efficiency advantage persists across all datasets and modalities. On video datasets like CDF and WD, \textbf{ConLLM} reduces inference time by more than 2x and uses less than half the memory compared to larger models like MiniGPT4-Video \cite{ataallah2024goldfish}. Similarly, in audio-visual tasks such as FAFC and DFDC, \textbf{ConLLM} demonstrates substantial reductions in computational costs while maintaining competitive performance.

\textbf{\textit{Note:}} We focus our comparison on domain-relevant multimodal models with native capabilities to handle audio, video, or audio-visual embeddings, deliberately excluding unimodal or purely text-based LLMs (e.g., GPT-3.5 or LLaMA) which are not directly comparable in this context.

\begin{figure}[t!]
    \centering
    \includegraphics[width=\linewidth]{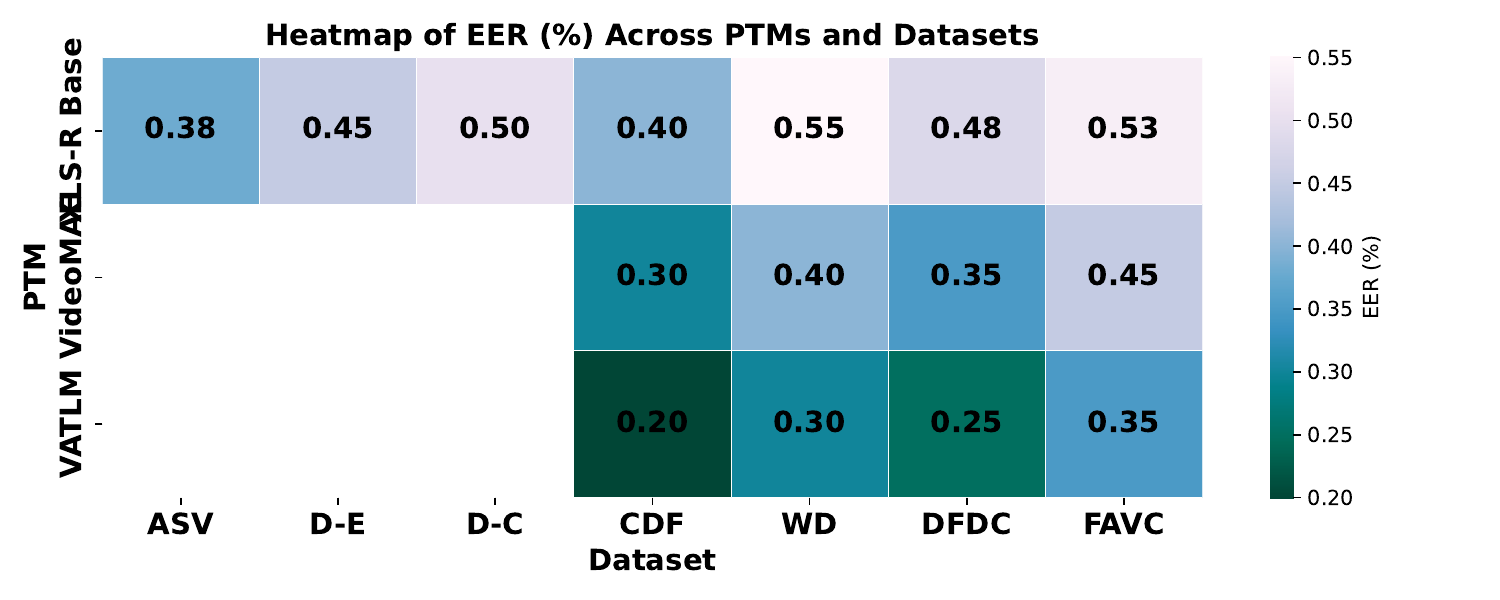}
    \caption{Heatmap of Equal Error Rate (EER \%) across different PTMs and datasets}
    \label{fig:eer_heatmap}
\end{figure}

\begin{figure*}[!t]
  \centering

  \begin{subfigure}[b]{0.92\textwidth}
    \centering
    \includegraphics[width=\linewidth]{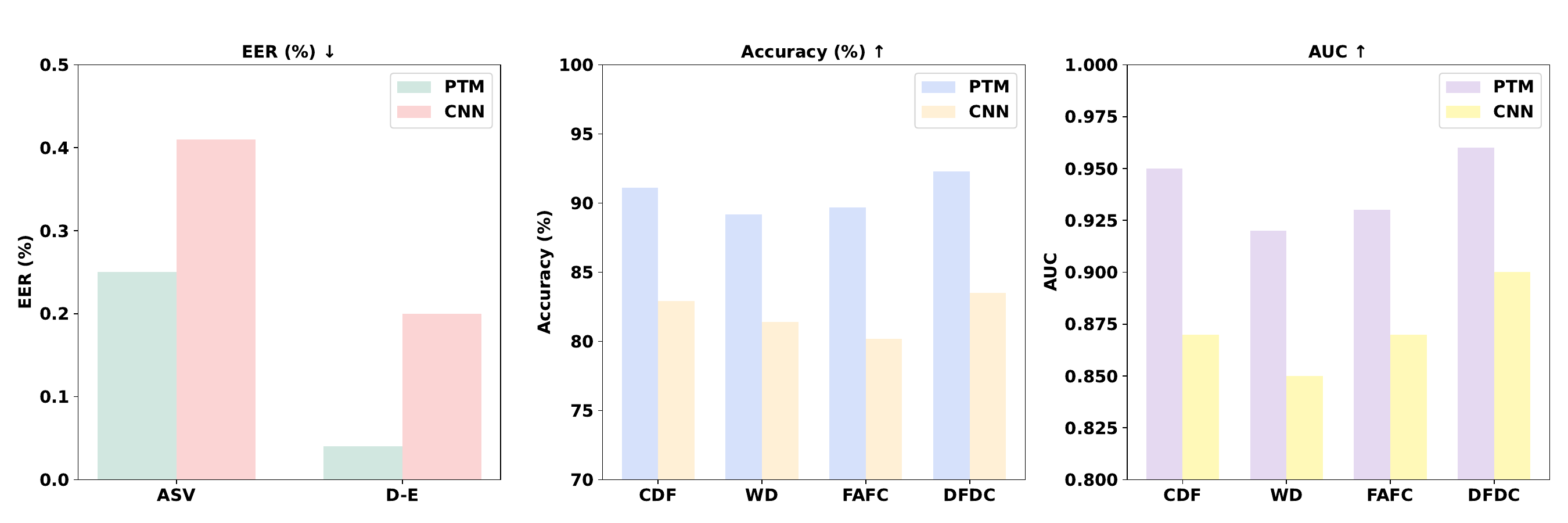}
    \caption{Performance Comparison of PTM-based and CNN-based Features}
    \label{subfig1}
  \end{subfigure}
  \hfill
  \begin{subfigure}[b]{0.92\textwidth}
    \centering
    \includegraphics[width=\linewidth]{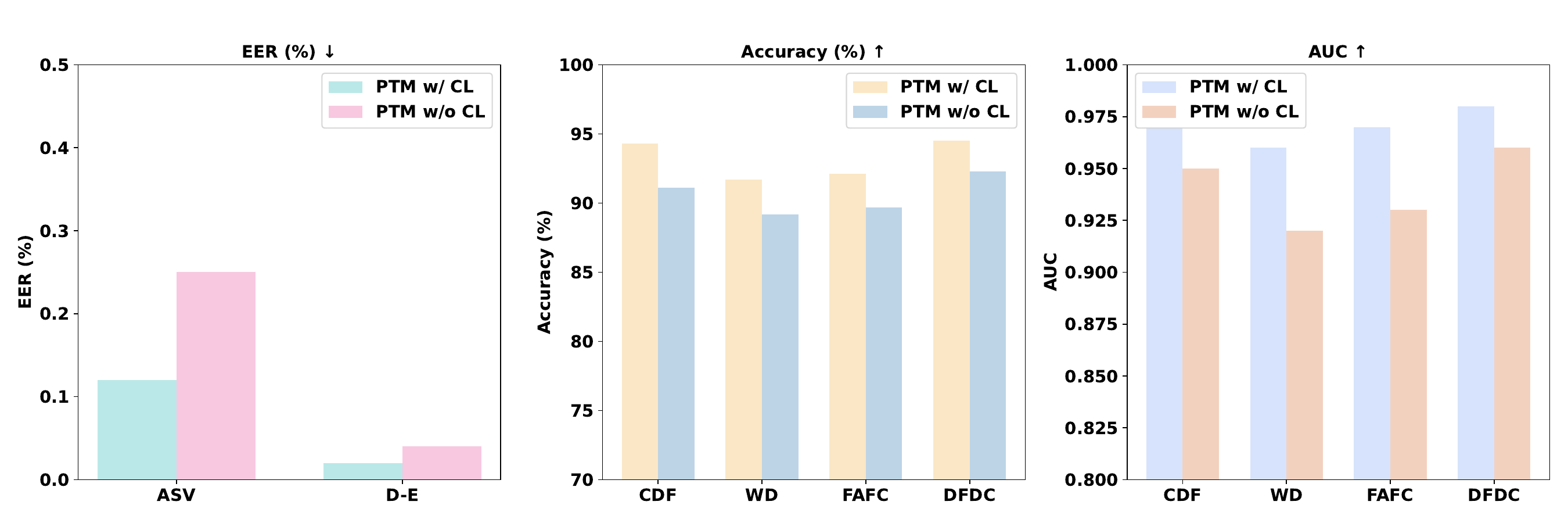}
    \caption{Performance Comparison of PTM with and without Contrastive Learning}
    \label{subfig2}
  \end{subfigure}

  \vskip 1em

  \begin{subfigure}[b]{0.92\textwidth}
    \centering
    \includegraphics[width=\linewidth]{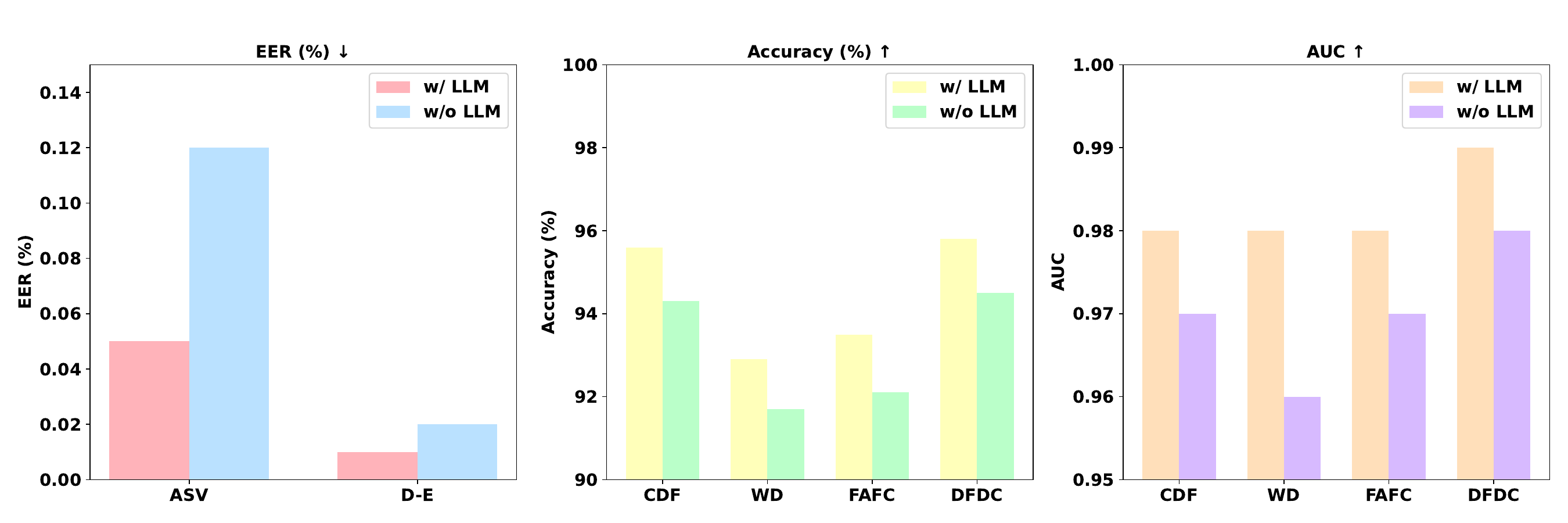}
    \caption{Performance Comparison of PTM with Contrastive Learning and LLM-Based Embedding Refinement vs. PTM with Contrastive Learning}
    \label{subfig3}
  \end{subfigure}
  \hfill
  \begin{subfigure}[b]{0.92\textwidth}
    \centering
    \includegraphics[width=\linewidth]{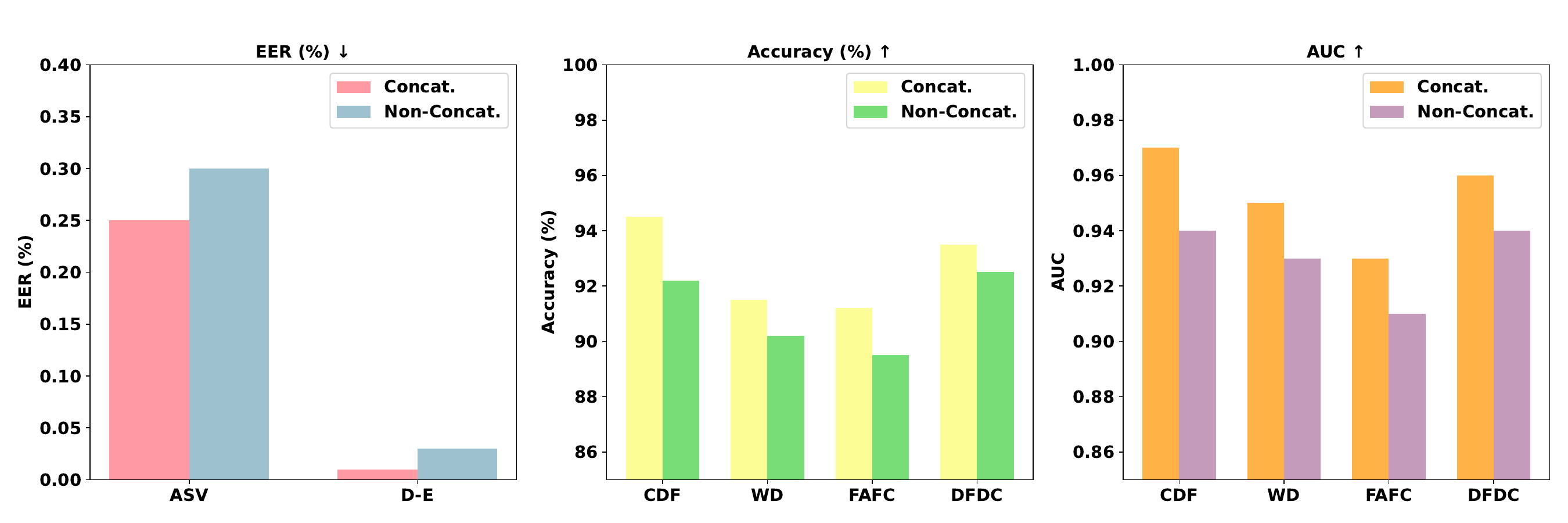}
    \caption{Performance Comparison of Fusion Strategies with PTM, Contrastive Learning, and LLM-Based Embedding Refinement}
    \label{subfig4}
  \end{subfigure}

  \caption{Ablations to evaluate the impact of (i) modality-specific embedding extraction using PTMs, (ii) contrastive learning for cross-modal alignment, (iii) embedding refinement via LLMs, and (iv) comparison to alternative fusion strategies}
  \label{fig:four_model_comparison}
\end{figure*}

\subsection{Additional Analysis}

To assess the performance of various PTMs in detecting deepfakes across multiple datasets, we evaluate the EER for each model, as shown in Figure \ref{fig:eer_heatmap}. The XLS-R Base model exhibits the most consistent performance, with EER values ranging from 0.38\% to 0.55\% across all datasets. VideoMAE performs well on the CDF, WD, DFDC, and FAFC datasets, achieving EERs between 0.30\% and 0.45\%, though its results are not available for the ASV, D-E, and D-C datasets. VATLM shows the lowest performance across datasets, with EER values between 0.20\% and 0.35\%, indicating comparatively limited effectiveness in these tasks.

\begin{table}[t!]
\scriptsize
\centering
\resizebox{\columnwidth}{!}{
\begin{tabularx}{\columnwidth}{l *{2}{>{\centering\arraybackslash}X}}
\toprule
\textbf{Model} & \textbf{D-C (Train) $\rightarrow$ D-E (Test)} & \textbf{D-E (Train) $\rightarrow$ D-C (Test)} \\
\midrule
XLS-R + x-vector & 16.14 & 35.72 \\
Whisper + Unispeech-SAT & 15.74 & 44.26 \\
\midrule
\textbf{ConLLM} & \textbf{1.5} & \textbf{2.0} \\
\bottomrule
\end{tabularx}
}
\caption{EER (\%)~\textdownarrow~for for Cross-Lingual Generalization against \cite{phukan2024heterogeneity}.}
\label{tableADAMANTFlexCrossLingual}
\end{table}

Furthermore, Table \ref{tableADAMANTFlexCrossLingual} compares the cross-lingual generalization of \textbf{ConLLM} on the DECRO dataset against \cite{phukan2024heterogeneity}, demonstrating superior generalization. Specifically, \textbf{ConLLM} achieves relatively low EERs when trained on the D-C dataset and tested on the D-E dataset (1.5\%), and vice versa, when trained on D-E and tested on D-C (2.0\%). These results indicate that \textbf{ConLLM} possesses strong cross-lingual generalization capabilities, effectively adapting across different language domains. The small variation in performance between the two transfer tasks highlights the model’s robustness in handling new linguistic contexts while maintaining low error rates in both directions.

\section{Ablation Studies}
\label{Ablation Studies}

We perform comprehensive ablation studies to analyze the contribution of key components in the \textbf{ConLLM} framework: (i) modality-specific embedding extraction using PTMs, (ii) contrastive learning for cross-modal alignment, (iii) LLM-based embedding refinement, and (iv) fusion strategies.

\paragraph{Modality-Specific Embedding Extraction.}  
Replacing PTM-based embeddings with traditional CNN features significantly degrades performance across datasets (see Fig.~\ref{subfig1}). On the ASV dataset, focused on voice spoof detection, EER increases from 0.21\% to 0.41\%, indicating CNNs’ limited capacity to capture subtle speech inconsistencies. Similarly, for D-E, EER rises from 0.04\% to 0.20\%. Video datasets also show performance drops; on CDF, accuracy decreases from 91.1\% to 82.9\% and AUC from 0.95 to 0.87, highlighting PTMs’ superior ability to capture fine-grained audio-visual cues. WD experiences a similar decline, with accuracy dropping from 89.2\% to 81.4\% and AUC from 0.92 to 0.85. For multimodal datasets FAFC and DFDC, PTMs yield notable gains in accuracy (80.2\% to 89.7\% on FAFC, 83.5\% to 92.3\% on DFDC) and AUC, underscoring their effectiveness in extracting rich multimodal representations. These results arise because PTMs are pretrained on large-scale diverse data and thus encode richer, more abstract, and context-aware features tailored for speech and visual modalities. Traditional CNNs, while effective for generic image tasks, lack the temporal and semantic understanding needed to detect subtle deepfake artifacts, leading to weaker discriminative power.

\paragraph{Contrastive Learning for Cross-Modal Alignment.}  
Introducing contrastive learning improves alignment of audio-visual embeddings (see Fig.~\ref{subfig2}). EER on ASV halves from 0.25\% to 0.12\%, while on D-E it improves from 0.04\% to 0.02\%. Video datasets benefit notably--CDF accuracy rises from 91.1\% to 94.3\% and AUC from 0.95 to 0.97; WD accuracy improves from 89.2\% to 91.7\%, AUC from 0.92 to 0.96. Multimodal datasets also show marked gains, with FAFC accuracy increasing from 89.7\% to 92.1\%, AUC from 0.93 to 0.97, and DFDC accuracy from 92.3\% to 94.5\%, AUC from 0.96 to 0.98. The observed improvements are due to contrastive learning’s ability to explicitly align embeddings from different modalities by bringing matching pairs closer and pushing mismatched pairs apart in the latent space. This alignment enhances the model’s sensitivity to cross-modal inconsistencies, which are critical cues for deepfake detection.

\paragraph{LLM-Based Embedding Refinement.}  
Further refinement with LLMs yields consistent improvements by incorporating semantic context (see Fig.~\ref{subfig3}). On ASV, EER improves from 0.12\% to 0.05\%, and on D-E from 0.02\% to 0.01\%. Video datasets like CDF see accuracy increase from 94.3\% to 95.6\%, AUC from 0.97 to 0.98, while WD accuracy rises from 91.7\% to 92.9\% and AUC from 0.96 to 0.98. For FAFC and DFDC, LLM refinement enhances accuracy to 93.5\% and 95.8\%, and AUC to 0.98 and 0.99 respectively. These gains result from the LLM’s capacity to incorporate high-level semantic understanding and contextual reasoning into the embeddings. Leveraging LLMs enables the model to capture subtle linguistic and contextual nuances that signal sophisticated forgeries, thereby enhancing detection robustness.

\paragraph{Fusion Strategies.}  
Comparing concatenation-based and non-concatenation fusion methods (see Fig.~\ref{subfig4}), concatenation consistently outperforms across datasets. On ASV, EER is 0.21\% versus 0.30\%; CDF accuracy is 94.5\% versus 92.2\%. WD accuracy is 91.5\% compared to 90.2\%, and FAFC and DFDC also show superior accuracy and AUC with concatenation fusion. This advantage arises because concatenation preserves modality-specific information by retaining full feature representations from each modality before joint processing, allowing richer joint reasoning. Non-concatenation fusion methods such as addition or averaging may dilute or lose important modality-specific signals early, leading to less discriminative combined embeddings.

\section{Conclusion and Future Works}

In this study, we proposed \textbf{ConLLM}, a state-of-the-art framework for deepfake detection that effectively leverages multimodal features and advanced learning techniques to deliver exceptional performance across diverse datasets. Our extensive experiments demonstrate that \textbf{ConLLM} consistently outperforms existing deepfake detection models, excelling in both unimodal and multimodal scenarios. This adaptability enables robust detection of emerging deepfake generation methods, highlighting its practical applicability in real-world settings.


\section*{Limitations}
\label{sec:Limitations}

Despite the promising performance of \textbf{ConLLM}, several limitations should be acknowledged. First, the model's performance depends heavily on the quality and diversity of the training data, which may limit its generalizability to novel or unseen deepfake generation techniques. Moreover, the computational demands for training and inference in multimodal deepfake detection pose challenges for deployment in real-time and resource-constrained environments. Finally, although \textbf{ConLLM} shows strong results on several popular benchmarks, further evaluation on a wider variety of real-world datasets is necessary to comprehensively assess its robustness, scalability, and interpretability.

\section*{Ethics Statement}
\label{sec:Ethics Statement}

The development and deployment of \textbf{ConLLM} are guided by ethical considerations to ensure responsible use in deepfake detection. The model aims to mitigate the risks associated with the proliferation of manipulated content, supporting efforts to combat misinformation and protect individuals from potential harm. However, it is important to acknowledge that the application of deepfake detection technologies raises concerns about privacy, consent, and potential misuse. To address these, we advocate for the transparent and ethical use of \textbf{ConLLM} while ensuring that appropriate safeguards are in place to protect personal and sensitive information.

\bibliography{main}

\end{document}